# Error Propagation in Satellite Multi-image Geometry

Joseph L Mundy and Hank Theiss

*Abstract*— This paper describes an investigation of the source of geospatial error in digital surface models (DSMs) constructed from multiple satellite images. In this study the uncertainty in surface geometry is separated into two spatial components; global error that affects the absolute position of the surface, and local error that varies from surface point to surface point. The global error component is caused by inaccuracy in the satellite imaging process, mainly due to uncertainty in the satellite position and orientation (pose) during image collection. The key sources of local error are; lack of surface appearance texture, shadows and occlusion. These conditions prevent successful matches between corresponding points in the images of a stereo pair.

A key result of the investigation is a new algorithm for determining the absolute geoposition of the DSM that takes into account the pose covariance of each satellite during image collection. This covariance information is used to weigh the evidence from each image in the computation of the global position of the DSM. The use of covariance information significantly decreases the overall uncertainty in global position and also results in a 3-d covariance matrix for the global accuracy of the DSM. This covariance matrix defines a confidence ellipsoid within which the actual error must reside. Moreover, the absolute geoposition of each image is refined to the reduced uncertainty derived from the weighted evidence from the entire image set.

The paper also describes an approach to the prediction of local error in the DSM surface. The observed variance in surface position within a single stereo surface reconstruction defines the local horizontal error. The variance in the fused set of elevations from multiple stereo pairs at a single DSM location defines the local vertical error.

These accuracy predictions are compared to ground truth provided by LiDAR scans of the same geographic region of interest. The prediction of global and local error is compared to the actual errors for a number of geographic locations and mixes of satellite type. The predicted error bounds contain the observed errors according to the allowed percentage of outliers.

## I. INTRODUCTION

This paper is focused on the problem of predicting the expected geo-spatial accuracy of digital surface models (DSMs) generated from general satellite image archives without the need for special stereo pair collection. Traditionally, DSMs are generated from special stereo pair collections from a single imaging satellite platform, and each image is equipped with extensive metadata describing the expected sensor position and attitude error variance as well as error in the line by line image formation geometry. Given this data, DSM accuracy is analyzed by a process known as *error propagation* [1] where known error covariances are passed through the algorithmic stages of stereo geometry formation to predict the full $3 \times 3$ covariance matrix associated with 3-d coordinates of the DSM.

Unfortunately, this support data is not readily available in the case of commercial image archives and so traditional error propagation within and across multiple stereo pairs cannot be readily applied. Thus, the goal of this paper is to introduce new methods to estimate predicted accuracy in multi-image DSM products and demonstrate these methods on a heterogeneous mix of satellite platforms and for various scene contexts and geographic locations.

*A. Digital Surface Models from Satellite Images*

Since the beginning of space-based imaging, specially-collected stereo pairs have been required to produce high-quality DSMs due to limitations in stereo reconstruction algorithms and to enable optimization of absolute geo-positioning of the surface model. The stereo pair is formed by collecting two images in near proximity on the satellite orbit such that the rows of the two images are aligned. This alignment enables efficient search for corresponding points along rows of the stereo pair. The need for such special satellite collections necessarily limited the availability of DSMs due to collection tasking priorities that favor single images for reconnaissance and commercial land use analysis.

The situation changed dramatically with the discovery that any pair of satellite images with overlapping views and rational polynomial coefficient (RPC) projection metadata can be rectified into a stereo pair [2]. The result is that vast archives of satellite image data can be exploited to produce DSMs for many regions on the Earth surface and over an extensive range of time intervals. Moreover, images from different satellite platforms can be combined into a stereo pair further extending the population of images that can be used to generate surface models.

### B. Semi-global Matching

Another key advance that makes the generation of accurate and high-resolution DSMs from general image archives feasible was the development of the semi-global matching (SGM) stereo algorithm by Hirschmüller [3]. The SGM algorithm provides a dense reconstruction of 3-d geometry based on matching corresponding pixels in the left and right stereo images using a dynamic programming scheme. The SGM algorithm makes a global assessment of the cost for assigning corresponding pixels and thus errors in matching left and right image pixels are greatly reduced, while at the same time maintaining computational efficiency.

### C. Geo-registration

To produce consistent geometry from multiple stereo pairs it is necessary to geo-register the set of input images into a common spatial coordinate frame with a high degree of accuracy. The registration must be achieved with sub-pixel accuracy so that a 3-d point produced by one stereo pair is closely aligned with the same point produced by other pairs. It is necessary to carry out the alignment of all the images in the dataset at the same time in order to achieve consistent geometry in a procedure called bundle adjustment [4]. The bundle adjustment algorithm relies on corresponding image features to provide constraints to solve for the unknown alignment transformation of each image. A set of image correspondences across a subset of the input images corresponding to a single 3-d point is called a *track*. Bundle adjustment simultaneously determines the 3-d point coordinates and adjusts the projection models of the track images to achieve accurate co-registration.

### D. Multiple Stereo Pair Fusion

Finally, it is necessary to fuse together multiple sets of densely reconstructed 3-d points, generated from their respective multiple stereo pairs, in order to form a complete surface. No single pair can produce a complete surface model given the occlusion present in a 3-d scene. For example, in urban environments many viewpoints are required to observe all surfaces due to occlusion by tall buildings. Similarly, cast shadows in the scene observed by a single stereo pair produce areas of unknown geometry due to the failure to find matching appearance values between a shadowed and visible surface point. Hirschmüller suggested some fusion strategies in his original paper, and the topic has received much attention in the literature [5,6]. For the case of a digital surface model, multiple stereo pairs produce a set of elevation values for each position in the $x, y$ plane and fusion is the process of selecting a single elevation that represents the set. One common approach is to find the median elevation. The median has been shown to be robust to elevation outliers and exhibits smoothly varying elevations over locally planar surface regions.

## II. TYPES OF DIGITAL SURFACE MODEL ERRORS

A DSM can be considered to be a function, $z(x, y)$ where a horizontal position, $(x, y)$, maps to a single elevation value, $z$. The DSM can be represented in a number of different coordinate systems such as WGS84 or UTM. In the case of WGS84 $x$ and $y$ are in degrees, and in UTM in meters. A DSM of the University of California at San Diego is shown in Fig. 1. A common format of a DSM is in the form of a GEOTIFF image where the image pixel location $(i, j)$ maps to a geographic position such as longitude and latitude in the case of WGS84. The image has floating point pixel values indicating the elevation in meters. A close up of this image format is shown in Fig. 1b). In this example the pixels are $0.3 \times 0.3$ meters square.

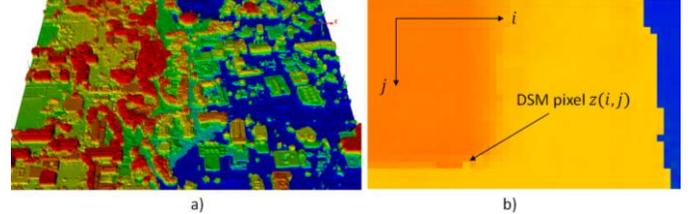

Fig. 1 a) A DSM for the campus of the University of California San Diego. The color coding indicates elevation values with red higher than blue. The total range is about 100m. b) A zoomed in view of a surface in a).

### A. Horizontal Error

Ideally, the $(i, j)$ location of each DSM pixel corresponds to the $(x, y)$ location of that surface position in the actual scene. The location will be quantized by the granularity of the DSM image pixel spacing, as in Fig. 1, to 0.3 meters. However, the error in the horizontal location can be much larger than the pixel spacing. Denote the actual ground truth location of pixel $(i, j)$ as $(x_{gt}, y_{gt})$ and denote the DSM position as specified by the GEOTIFF image as $(x_{dsm}, y_{dsm})$. The $(x, y)$ error is defined as,

$$\varepsilon_{Horizontal} = (x_{dsm} - x_{gt})\hat{x} + (y_{dsm} - y_{gt})\hat{y} \qquad (1)$$

typically called *horizontal* error since the error vector lies in the horizontal $x, y$ plane.

### B. Vertical Error

Even if the horizontal position is perfectly aligned with its ground truth position there can be error in the elevation, $z(x, y)$, specified by the DSM at each $(x, y)$ location. The DSM elevation is the result of fusing many stereo pair estimates for $z(x, y)$ but there can be outliers and so the fused result can be perturbed from its ground truth value. Moreover, each stereo pair may not be in exact georegistration and so there can be blurring of surface features due to fusing the displaced surfaces.

### C. Global Error

Another component of horizontal and vertical error is due to a global transformation of the entire DSM with respect to the correct geographic location of the surface. Such global transformations are caused by error in the image geo-registration process. The geo-registration algorithm does not make use of any known ground points but relies on the bundle-adjustment algorithm to average out errors in the metadata that defines the projection of a 3-d point into image coordinates. A large component of this type of error is due to uncertainty in the pointing direction of the satellite during image collection. The satellite is hundreds of kilometers from the Earth surface and so even a few micro-radians of pointing error can cause meters of positional error at the surface. If the satellite metadata is biased then a residual global 3-d error in the position of the entire DSM will occur.

*D. Local Error*

Horizontal and vertical errors can vary from pixel to pixel and become larger near the boundaries of building roofs or other sharp changes in surface elevation. These rapid spatial variations are due mainly to error in pixel correspondences within the SGM algorithm due to variations in appearance between images so that errors in stereo correspondence occur. For example, some locations can be missed entirely if one image in the stereo pair has a surface point in shadow and the other does not.

Additionally, error can arise due to the quantization of corresponding locations in a stereo pair since the matching precision is $\pm 1$ pixel in rectified image space, and depending on the view angle separation of the pair, this quantization can translate to a large error in the 3-d position of the intersection point. That is, the geometric accuracy of the intersection of two rays degrades as the rays become more parallel.

Another source of horizontal error is the assignment of 3-d points generated from the stereo pairs to quantized DSM pixel locations. The 3-d points assigned to a pixel bin form the population that is used to fuse the $z$ value for that bin. However, the centroid of the population is not generally at the center of the pixel and so a shift in horizontal location for the fused $z$ value is generated.

*E. Previous approaches*

a)   Global error

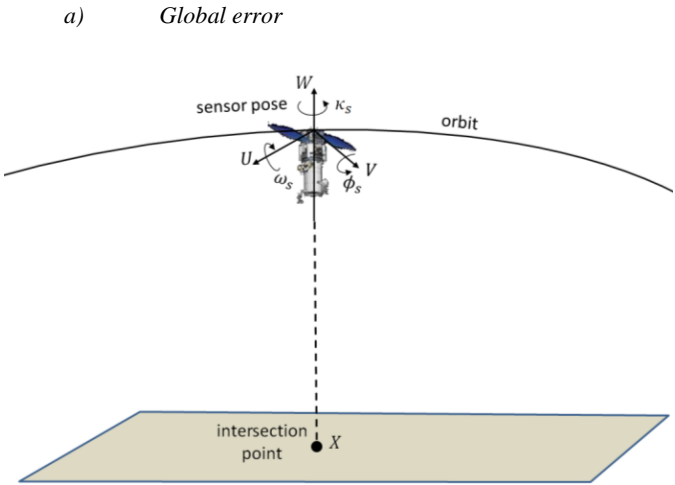

Fig 2 The imaging satellite pose covariance coordinate system. The covariance of intersection point X is derived from the known covariance of sensor pose.

A standard approach to model global error of multi-image geopositioning (MIG) such as a DSM is to propagate known error covariance in the pose (position and attitude) of an imaging satellite to the $3 \times 3$ covariance matrix of 3-d point coordinates [7]. The coordinate system for satellite position and attitude (pose) error is illustrated in Fig 2. The analysis for the covariance of $X$ for $n$ satellite images requires three matrices: $\Sigma_p$ the $6n \times 6n$ covariance matrix for satellite pose errors; $B_p$ the $2n \times 6n$ Jacobian matrix of derivatives of image coordinates with respect to pose; and $B$ the $2n \times 3$ matrix of derivatives of image coordinates with respect to the coordinates of $X$.

Denote the error in the intersection point as $\partial X$ so that $X = X_0 + \partial X$. The covariance matrix for $\partial X$ is denoted as $P = \langle \partial X \partial X^T \rangle$, where $\langle \cdot \rangle$ indicates expectation over an ensemble of random instances. Given these definitions the error propagation proceeds as follows.

$$W = (B_p \Sigma_p B_p^T)^{-1} \qquad (2)$$
$$P = (B^T W B)^{-1} \qquad (3)$$

The $2n \times 2n$ weight matrix $W$ is the inverse covariance matrix for errors in image coordinates due to satellite pose errors. This weight matrix propagates pose errors to produce the covariance matrix $P$ for intersection point coordinates via the matrix $B$. This analysis in terms of derivatives is valid if the intersection point $X$ is near its true value and the pose errors are small enough so that first derivatives of the projection from 3-d to 2-d are sufficiently accurate.

There are two limitations of this approach: 1) it is necessary to have an initial guess for the intersection point; 2) there is no prescription for the application of satellite pose uncertainty to the computation of the initial guess for the intersection point. Instead, a different procedure to produce the initial guess is recommended such as least-squares line intersection [8]. The method does however provide a way of refining the initial guess based on the pose covariance information. The value of $X$ can be iteratively updated by,

$$\partial X = P B^T W (\tilde{x} - x), \qquad (4)$$

where $x$ is the image projection of $X$ and $\tilde{x}$ is the actual image position of $X$.

b)   Local error

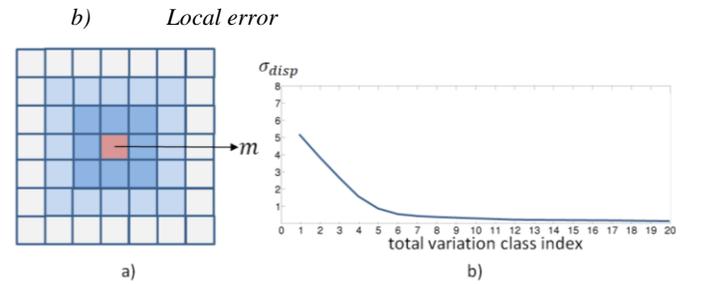

Fig 3 Estimating local errors in disparity. a) The set of neighborhood rings for computing total variation class. b) The relation between total variation class and the standard deviation of disparity.

There have been previous efforts to model the effect of the types of local errors defined above. A method called *total variation* can be applied to the disparity image generated by the SGM algorithm. The disparity image encodes the row by row pixel shift (disparity) required to match each pixel in the one stereo pair image to a pixel on the same row in the other image. It has been proposed by Kuhn that localized, large changes in disparity will be characteristic of high uncertainty in the reconstructed 3-d geometry [9]. Kuhn calls the quantification of these rapid changes total variation.

The computation of total variation is illustrated in Fig 3a). The class index of total variation of disparity is computed as,

$$TV_{class} = \max_n \left( \sum_{m=1}^{n} \frac{1}{8m} \sum_{i,j \in N_m} \sqrt{|d_{i+1,j} - d_{i,j}| + |d_{i,j+1} - d_{i,j}|} < \theta \right) \qquad (5)$$

, where $m$ is the radius of a neighborhood of pixels, $N_m$ as shown in Fig 3a), and $d_{i,j}$ is the disparity at neighborhood location $(i, j)$. If the disparity image is smooth, a large radius is required before changes in disparity occur that exceed the threshold $\theta$. Conversely, if the disparity region is noisy then the threshold will be exceeded for small values of $m$. The index value can be mapped to the corresponding value of actual standard deviation in disparity by observing the values of $TV_{class}$ on training data equipped with ground truth disparity values. An example curve relating $TV_{class}$ to the standard deviation of disparity, $\sigma_{disp}$, is shown in Fig 3 b). The standard deviation in disparity is then used to compute the covariance matrix of the reconstructed 3-d points for the stereo pair.

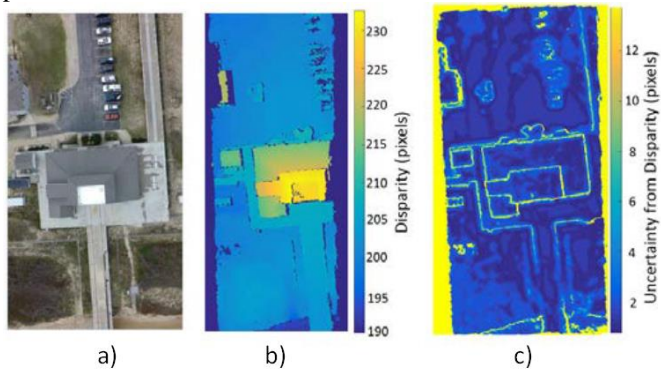

Fig. 4 a) Drone image. b) Disparity image. c) Predicted uncertainty in disparity. From Rodarmel *et al* [10].

This approach has been applied to estimating the local errors in 3-d point clouds derived from aerial images collected by a small drone. The work by Rodarmel *et al*. [10] applies rigorous error propagation methods to predict spatially varying point covariance. An example of predicted disparity standard deviations from the disparity image, see Fig. 4c), is shown to bound the actual errors as determined from ground truth point positions.

*F. Proposed approach*

A new approach that is keyed to the specific case of DSM construction from satellite image is proposed. In the development to follow, a typical multi-image stereo DSM reconstruction system is used to illustrate the method and study the results of applying satellite pose covariance to the prediction of local and global errors manifested for a given mix of satellite sensor error characteristics. The DSM reconstruction computational pipeline is formulated in terms of services such as radiometric correction, region of interest tiling, image geo-registration, stereo-pair processing and fusion, hereafter known as the "system."

*a)    Global error*

The prediction of global accuracy is based on propagating the pose covariance of the satellite sensors throughout the geo-registration process and its construction of the 3-d intersection point for matched feature correspondences across multiple images (called a track). Such tracks are found by matching features such as SIFT [11] between pairs of images and recursively building up matches among a set of images. The track feature correspondences enable the simultaneous determination of the 3-d point formed by the intersection of rays back-projected from each feature location and the image translations required to co-register the 3-d to 2-d projection functions for the track images.

In this approach, the satellite pose covariance informs the determination of the 3-d track point by weighing the accuracy of sensor rays cast from each track correspondence. The ray intersection algorithm uses covariance-weighted least squares to find the 3-d point that minimizes the sum of perpendicular distances from the rays to the point. The predicted covariance of the 3-d track point emerges as an integral result of the intersection algorithm. Global DSM placement is rigidly attached to the 3-d track point since the images are all co-registered with respect to the forward projection of the point into each image. The covariance of the global DSM position as a whole then corresponds to the covariance in the location of the 3-d track point. Since the solution is based on linear least squares, the result is obtained in closed form not requiring an iterative solution with the attendant risk of non-convergence.

*b)    Local error*

The prediction of local vertical and horizontal accuracy is based on an analysis of the fusion of stereo point clouds. A stereo pair produces a dense point cloud with a point generated for each valid location in the disparity map. A typical stereo pair will generate millions of points, some of which are invalid due to incorrect correspondences found by the semi-global matching (SGM) algorithm. In the proposed approach the SGM algorithm is run twice, a second time with the images processed in reverse order. The 3-d points generated by the forward and reverse order of the stereo pair are compared and a high probability assigned to a 3-d point if its forward and reverse locations are close to each other and lower probabilities as the distance between the forward and reverse locations increases.

Unlike the comparison of disparity values, geometric distance comparison is invariant to stereo pair view separation and so the point probability values for different pairs can be compared. The probabilities are used to weigh the evidence from multiple stereo pairs in forming the pixel bin populations and in determining the elevation value for each bin. This weighted geometric information can then be used to compute the horizontal and vertical variances for each DSM pixel location, thus taking into account the disparity errors of the SGM algorithm and error in 3-d location based on disparity. Typically 100s of pairs are combined to form the final DSM surface but as few as three pointsets are sufficient to compute horizontal and vertical variances, albeit with low accuracy.

The approach just described provides for local and global error prediction and presents a unified approach based on the geometric uncertainty of rays cast from each sensor pixel into 3-d space. The formulation in terms of rays is agnostic to the type of sensor and so the analysis can apply to a wide range of overhead imaging systems including complex wide area motion image (WAMI) scanners and single perspective focal plane cameras as well as a heterogeneous constellation of satellite imaging systems, which is the focus of this paper.

## III. THE RAY MODEL FOR SATELLITE IMAGES

### A. Geometric Rays

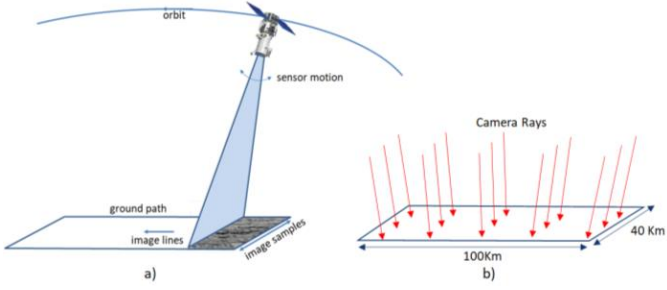

Fig. 5 The formation of a satellite image. a) The image is built up line-by-line. b) The collection of ray cast from image pixels as seen at the corresponding ground points.

The formation of a satellite image is shown in Fig. 5a). The satellite collects the image lines through a combination of orbital and satellite scanning motion. The projection from ground space to image space is determined from the position and attitude of the spacecraft for each line of the image, as well as the optical system geometry. Any error in the determination of satellite pose will result in an error in the geographic location of the camera rays and thus in the location of the reconstructed geometry from multiple satellite images. Given a satellite orbit 500Km above the Earth surface, small attitude errors manifest themselves as translations of the image with respect to the actual ground location. Thus correction of these pointing errors during geo-registration can be achieved by a 2-d translation of the image.

As shown in Fig. 5b), a typical commercial satellite image spans a region that is 10s of kilometers in extent. An example set of rays is shown in the figure where it can be seen that the ray directions are locally nearly parallel but gradually change direction as the satellite moves along its orbit. The ray geometry, origin and direction, can be derived directly from the third order rational polynomial coefficient (RPC) metadata by computing the back-projection of a pixel location onto a pair of 3-d planes displaced in elevation. The resulting pair of points defines the ray geometry. However, the non-linear back-projection computation is costly if carried out at every pixel in the image.

### B. The affine camera

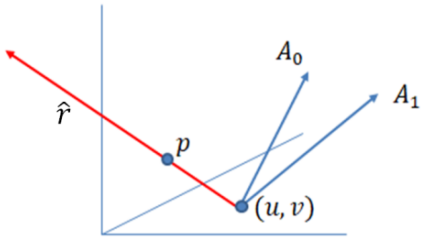

Fig. 6 The ray $r$ defined by the image location (u, v) for an affine camera. $\hat{r}$ is the unit ray direction vector and $p$ is an arbitrary point on the ray. $A_0$ and $A_1$ are 3-d row vectors from the upper $2 \times 3$ sub-matrix of the affine camera matrix. The ray direction is defined by $A_0 \times A_1$.

The computational cost of reconstructing rays can be considerably reduced by segmenting the region of interest into rectangular tiles that encompass a small enough region that the rays can be considered to be parallel. This condition is equivalent to an affine camera model [12]. The RPC metadata can be used to construct the affine camera parameters for a given tile by randomly generating a set of 3-d to 2-d correspondences using the forward projection of the RPC model. The affine camera parameters are then estimated from the correspondence set using a linear regression algorithm. The affine camera projection is defined by a $3 \times 4$ matrix,

$$C_{aff} = \begin{bmatrix} a_{00} & a_{01} & a_{02} & a_{03} \\ a_{10} & a_{11} & a_{12} & a_{13} \\ 0 & 0 & 0 & 1 \end{bmatrix} = \begin{bmatrix} A_0 & a_{03} \\ A_1 & a_{13} \\ 0 & 1 \end{bmatrix} \quad (6)$$

, so that

$$\begin{bmatrix} u \\ v \\ 1 \end{bmatrix} = \begin{bmatrix} A_0 & a_{03} \\ A_1 & a_{13} \\ 0 & 1 \end{bmatrix} \begin{bmatrix} X \\ Y \\ Z \\ 1 \end{bmatrix} \text{ or,} \quad (7)$$

$$\begin{bmatrix} u \\ v \end{bmatrix} = \begin{bmatrix} A_0 \\ A_1 \end{bmatrix} \begin{bmatrix} X \\ Y \\ Z \end{bmatrix} + \begin{bmatrix} a_{03} \\ a_{13} \end{bmatrix} \quad (8)$$

As shown in Fig. 6, the ray direction $\hat{r}$ is given by the cross product of $A_0$ and $A_1$ treated as vectors. The direction $\hat{r}$ is the same for any pixel in the image region for which the affine camera is valid. The definition of the ray is completed by specification of any point $p$ along the ray as shown in Fig. 6. Given an image location $(u, v)$, the point is given by $p = \beta_0 A_0 + \beta_1 A_1$, where

$$\begin{bmatrix} \beta_0 \\ \beta_1 \end{bmatrix} = \begin{bmatrix} A_0 \cdot A_0 & A_0 \cdot A_1 \\ A_0 \cdot A_1 & A_1 \cdot A_1 \end{bmatrix}^{-1} \begin{bmatrix} u - a_{03} \\ v - a_{13} \end{bmatrix} \quad (9)$$

All of these matrix operations can be carried out with good computational efficiency within a tile. For example the matrix inverse in Eq. (9) is the same for every pixel $(u, v)$ in the image region over the tile and so only needs to be computed once.

## IV. SATELLITE POSE COVARIANCE

### A. Satellite coordinate systems

There are a number of coordinate systems involved in computing the pose covariance matrix and its effect on the intersection of sensor rays in the formation of multi-image geometry, as shown in Fig. 7. The relationship between the satellite sensor and a local coordinate system on the ground is shown in a). A tangent plane to the Earth surface is defined at an origin point, $R_o$, shown in yellow in the figure. An East-North-up (enu) coordinate system is constructed at $R_o$. The North ($N_o$) and up ($up_o$) axes are shown in the figure. The orientation of the satellite with respect to the enu coordinate system is defined by the azimuth and elevation angles shown in the figure. A unit vector $\hat{u}_s$ is computed from the azimuth and elevation angles and is expressed in the Earth-centered, Earth-fixed (ecf) coordinate system. The position of the satellite $R_s$ is computed from the known height of the orbit $H_s$ and the nominal radius of the Earth $R_e$.

Fig. 7b) shows the in-track, cross-track, radial (icr) coordinate system. The radial vector corresponds to $\hat{u}_s$. The tangent vector to the orbit (in-track) is computed from the known inclination angle, $\theta$, of the ground track of the satellite. The cross track coordinate vector is defined by the cross product of the radial and in-track vectors.

Fig. 7c) defines the sensor coordinate system. A satellite can change its orientation during image collection and so the sensor coordinate system is generally not aligned with the icr coordinate system. The sensor coordinates $U, V, W$ are defined such that $W$ is aligned with the view direction of the optical system, and pointing away from the Earth. The $V$ axis is perpendicular to $W$ and parallel to the scan direction of the image. The unit vector $U$ is defined as the cross-product of $V$ and $W$, which completes the sensor coordinate system. The angles $\omega, \phi, \kappa$ represent variations with respect to the measured attitude of the satellite and are a key source of pointing inaccuracy.

Satellite imagery is typically accompanied by metadata that specifies the following information:
- rational polynomial coefficient (RPC) model for projecting ground points to image points
- the origin point, $R_o$ on the Earth surface in WGS84 coordinates, $(\lambda_o, \phi_o)$.
- satellite azimuth $az$ and elevation $eE$ during image collection
- altitude of the satellite orbit, $H_s$
- inclination angle $\theta$ of the orbit ground track with respect to the equator

In addition to this information there are a number of geographic constants required such as the average radius of the Earth, $R_e = 6371000$m and the major and minor radii denoted as $a = 6378137.0$ m and $b = 6356752.31424518$ m respectively. With this information, $R_o$ can be transformed to Earth-centered-Earth-fixed (ecf) coordinates. The following sections describe the computation of the various coordinate frames required to determine the covariance of satellite pose errors.

## B. Determining satellite position

The location of the satellite $R_s$ in the ecf coordinate system is determined as follows. As shown in Fig. 7 a), the radius of the satellite orbit from the Earth center is $R_t = R_e + H_s$. The unit vector pointing from the origin to the satellite position in enu coordinates is

$$\hat{u}_s = \begin{bmatrix} Cos(eE)Sin(az) \\ Cos(eE)Cos(az) \\ Sin(eE) \end{bmatrix}. \quad (10)$$

Note that azimuth is defined clockwise from North and elevation as zero on the tangent plane at $R_o$. In this convention a nadir image has eE = 90°. $\hat{u}_s$ is transformed to Earth-centered-Earth-fixed (ecf) coordinates as $\hat{u}_o = T_{enu \to ecf} \hat{u}_s$, where

$$T_{enu \to ecf} = \begin{bmatrix} -Sin(\lambda_o) & -Sin(\phi_o)Cos(\lambda_o) & Cos(\phi_o)Cos(\lambda_o) \\ Cos(\lambda_o) & -Sin(\phi_o)Sin(\lambda_o) & Cos(\phi_o)Sin(\lambda_o) \\ 0 & Cos(\phi_o) & Sin(\phi_o) \end{bmatrix}. \quad (11)$$

A sphere of radius $R_t$ is intersected with the ray from $R_o$ with direction $\hat{u}_o$ pointing towards the satellite. That is, the location $R_s$ of the satellite in the ecf coordinate frame can be computed as $R_s = R_o + k\,\hat{u}_o$ must lie on a sphere with radius $R_t$. The solution for the scale factor is

$$k = -R_o \cdot \hat{u}_o + \sqrt{(R_o \cdot \hat{u}_o)^2 - (R_o \cdot R_o - R_t^2)}. \quad (12)$$

The satellite's icr orbital coordinate frame is determined from the radial vector $R_s$ just computed and from the inclination angle of the orbit ground track. As an example, the WorldView3 satellite has an inclination angle of 97.7783°. Sun-synchronous satellites collect images on the descending pass and so the counter-clockwise ground track angle with respect to East is $\theta = 360 - 97.7783 = 262.2°$. The unit direction vector of the ground track in enu coordinates is $V_s^{enu} = \begin{bmatrix} Cos\,\theta \\ Sin\,\theta \\ 0 \end{bmatrix}$. This vector is transformed to ecf coordinates,

$$V_s^{ecf} = T_{s\,enu \to ecf} V_s^{enu}$$

, where $T_{s\,enu \to ecf}$ is obtained as follows. Define $z_u = \frac{R_s}{|R_s|}$, $x_u = \frac{\hat{z}_{ecf} \times z_u}{|\hat{z}_{ecf} \times z_u|}$, $y_u = z_u \times x_u$, then $T_{s\,enu \to ecf} = [x_u \quad y_u \quad z_u]$, a $3 \times 3$ rotation matrix with $x_u$, $y_u$ and $z_u$ as column vectors. Note that in ecf coordinates $\hat{z}_{ecf} = \begin{bmatrix} 0 \\ 0 \\ 1 \end{bmatrix}$ the North Pole unit vector. The icr coordinate system is completed as follows.

$$\hat{\imath} = \frac{V_s^{ecf}}{|V_s^{ecf}|}$$
$$\hat{c} = \frac{R_s \times \hat{\imath}}{|R_s \times \hat{\imath}|}$$
$$\hat{r} = \hat{\imath} \times \hat{c}$$
$$T_{icr \to ecf} = [\hat{\imath} \quad \hat{c} \quad \hat{r}]$$

The error in satellite position $R_s$ is denoted as

$$\mathcal{E}_{pos} = \begin{bmatrix} dI \\ dC \\ dR \end{bmatrix}.$$

The $3 \times 3$ covariance matrix $S_{pos} = \langle \mathcal{E}_{pos} \mathcal{E}_{pos}^T \rangle$ is assumed diagonal with equal variance,

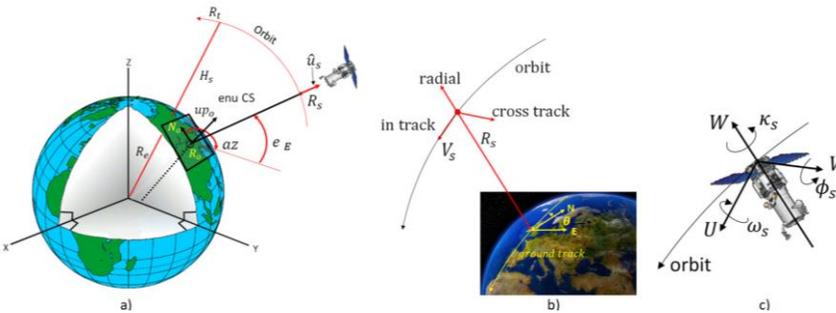

Fig. 7 Coordinate systems involved in specifying pose covariance. a) The position of the satellite in East-North-Up (enu) coordinates. b) The in-track, cross-track, radial (icr) orbital coordinate frame. c) The sensor coordinate frame. The angles $\omega, \phi, \kappa$ are relative to the measured attitude of the satellite.

$$S_{pos} = \begin{bmatrix} \sigma_{pos}^2 & & \\ & \sigma_{pos}^2 & \\ & & \sigma_{pos}^2 \end{bmatrix} \quad (13)$$

For example, the position variance for WorldView3 is specified as $\sigma_{pos}^2 = 0.5m^2$. This positional variance is obtained by comparing the predicted position of a point on the ground from its image location with the known position. There are numerous such ground control points that can be used to compute satellite position and attitude error variances. The 3-d point position is typically determined by ground survey methods such as differential GPS.

*C. Satellite attitude*

An imaging satellite can manipulate the orientation (attitude) of the optical system so as to collect off-nadir views. This rotated frame is called the *sensor* coordinate system. The orientation can be determined from the RPC model by projecting the center column of the image onto the enu tangent plane. The angle the projected column makes with the East axis is defined as $S_\theta$. The scan unit vector in enu coordinates is $S_{enu} = \begin{bmatrix} Cos\ S_\theta \\ Sin\ S_\theta \\ 0 \end{bmatrix}$. A common scan direction is from North to South and so $S_{enu} = \begin{bmatrix} 0 \\ -1 \\ 0 \end{bmatrix}$. In ecf coordinates, $S_{ecf} = T_{enu \to ecf} S_{enu}$. The sensor coordinate axis unit vectors are defined as,

$$s\hat{Z} = \frac{(R_s - R_o)}{|R_s - R_o|}, \quad s\hat{Y} = \frac{s\hat{Z} \times S_{ecf}}{|s\hat{Z} \times S_{ecf}|}, \quad s\hat{X} = s\hat{Y} \times s\hat{Z}$$

A matrix $M$ is defined that expresses the rotation of the sensor coordinate system with respect to the ecf coordinate system, where,

$$M = \begin{bmatrix} s\hat{X}^T \\ s\hat{Y}^T \\ s\hat{Z}^T \end{bmatrix} \quad (14)$$

The rotation from the enu coordinate frame to the sensor coordinate system is given by $M_{enu} = M T_{enu \to ecf}^T$. The matrix $M_{enu}$ defines the nominal attitude of the sensor during image collection in the local Cartesian coordinate frame where DSM geometry is defined

*D. Pose covariance*

A physical basis for pose uncertainty is now defined with respect to errors in satellite position and orientation (pose), so-called *adjustable parameters*. The error in satellite pose is defined as,

$$d\wp = \begin{bmatrix} dI \\ dC \\ dR \\ \omega_s \\ \phi_s \\ \kappa_s \end{bmatrix}, \quad (14)$$

where $(dI, dC, dR)$ are the in-track, cross-track and radial errors in satellite position and $(\omega_s, \phi_s, \kappa_s)$ are small angle errors in satellite sensor attitude as shown in Fig. 7c). The position error vector can be transformed into the sensor coordinate frame by $M\ T_{icr \to ecf}$ so that

$$\begin{bmatrix} dU \\ dV \\ dW \end{bmatrix} = T_{ecf \to sensor} T_{icr \to ecf} \begin{bmatrix} dI \\ dC \\ dR \end{bmatrix}, \quad (15)$$

where $T_{ecf \to sensor}$ is the rotation from ecf coordinates to the sensor coordinate system $(sX, sY, sZ)$, and $T_{icr \to ecf}$ is the rotation from the satellite in-track, cross-track and radial coordinate system to ecf coordinates. See Fig. 7c). This transformation will be incorporated later in the Jacobian that maps the pose errors into errors in image ray displacements.

The pose covariance matrix, denoted as $S_\wp$, is a $6n \times 6n$ matrix, where $n$ is the number of images. For a single image and for the WorldView3 satellite, the covariance matrix is,

$$S_\wp = \begin{bmatrix} 0.5m^2 & & & & & \\ & 0.5m^2 & & & & \\ & & 0.5m^2 & & & \\ & & & 8 \times 10^{-12} rad^2 & & \\ & & & & 8 \times 10^{-12} rad^2 & \\ & & & & & 16 \times 10^{-12} rad^2 \end{bmatrix}$$

As an illustration, for a nadir image the corresponding variance in ground position is,

$$\sigma_{XX} = 620000^2 \times 8 \times 10^{-12} + 0.5 = 3.6m^2$$

where the orbital altitude is 620 km.

It is also possible for sensor pose errors of two or more images to be correlated. The correlation is due to the persistence of error states for short periods of time along the same orbital pass. Sensor position and attitude errors for nearby images are highly correlated; i.e., a correlation coefficient on the order of $\rho = 0.8$. Pose errors are typically modeled as completely uncorrelated between images acquired from separate orbital passes even if acquired from the same sensor.

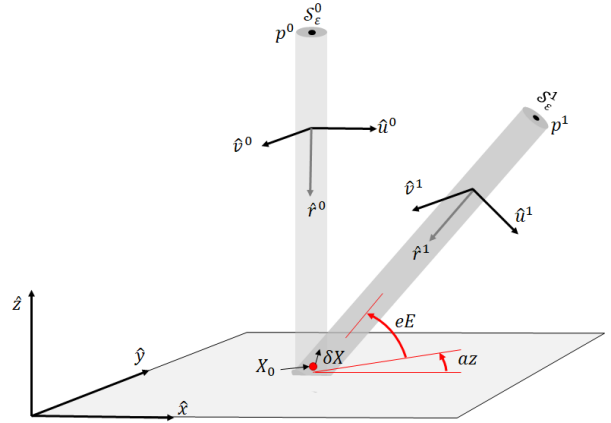

Fig. 8 The intersection of two rays. The ray direction unit vectors are $\hat{r}^0$ and $\hat{r}^1$ with origin points, $p^0$ and $p^1$, and covariance matrices $S_\varepsilon^0$ and $S_\varepsilon^1$ respectively.

V. PROPAGATING SATELLITE POSE COVARIANCE

*A. Ray covariance*

The intersection of two sensor rays is illustrated in Fig. 8. The sensor ray is shown as a semitransparent cylinder indicating that only displacements of a ray orthogonal to the ray affect the intersection point $X = X_0 + \partial X$. The ray displacements are due to pose errors during image collection. The covariance of ray origin $p^i$ is defined with respect to the unit vectors, $\hat{u}^i$ and $\hat{v}^i$ which are the unit sensor coordinate vectors $s\hat{X}$ and $s\hat{Y}$ transformed into the local enu Cartesian coordinate frame, denoted as $(\hat{x}, \hat{y}, \hat{z})$ in Fig. 8. $\hat{u}^i$ and $\hat{v}^i$ are orthogonal to each other and to the ray direction, $\hat{r}^i$. The perturbed ray origin is given by, $\tilde{p}^i = p^i + \varepsilon_u^i \hat{u}^i + \varepsilon_v^i \hat{v}^i$, where $(\varepsilon_u^i, \varepsilon_v^i)$ are random variables representing the ray displacement errors due to errors in sensor pose. For $n$ images the collective error vector is denoted as,

$$\mathcal{E}_s = \begin{bmatrix} \varepsilon_u^0 \\ \varepsilon_v^0 \\ \varepsilon_u^1 \\ \varepsilon_v^1 \\ \vdots \\ \varepsilon_u^{n-1} \\ \varepsilon_v^{n-1} \end{bmatrix} \quad (16)$$

and the ray covariance matrix $\mathcal{S}_{\varepsilon_s} = \langle \mathcal{E}_s \mathcal{E}_s^T \rangle$.

In order to relate the satellite pose error covariance to the ray covariance it is necessary to construct a Jacobian derivative matrix. For a single sensor the orthogonal ray displacements are,

$$\begin{bmatrix} \varepsilon_u \\ \varepsilon_v \end{bmatrix} = \begin{bmatrix} (dU(dI, dC, dR) + |R|\phi_s) \\ (dV(dI, dC, dR) - |R|\omega_s) \end{bmatrix} \quad (17)$$

Note that displacements along the ray and rotations about the ray, $\kappa_s$ do not affect the intersection point, $X$. The elements involving $\kappa_s$ of the pose covariance matrix, $\mathcal{S}_\wp$, can be dropped giving a $5n \times 5n$ matrix. The full $2n \times 5n$ Jacobian matrix is,

$$J = \frac{\partial \mathcal{E}_s}{\partial \wp} = \quad (18)$$

$$\begin{bmatrix} dU_I^0 & dU_C^0 & dU_R^0 & 0 & |R^0| & \cdots & 0 & 0 & 0 & 0 & 0 \\ dV_I^0 & dV_C^0 & dV_R^0 & -|R^0| & 0 & \cdots & 0 & 0 & 0 & 0 & 0 \\ \vdots & \vdots & \vdots & \vdots & \vdots & \ddots & \vdots & \vdots & \vdots & \vdots & \vdots \\ 0 & 0 & 0 & 0 & 0 & \cdots & dU_I^{n-1} & dU_C^{n-1} & dU_R^{n-1} & 0 & |R^{n-1}| \\ 0 & 0 & 0 & 0 & 0 & \cdots & dV_I^{n-1} & dV_C^{n-1} & dV_R^{n-1} & -|R^{n-1}| & 0 \end{bmatrix}$$

The $2n \times 2n$ ray covariance matrix is then $\mathcal{S}_{\varepsilon_s} = J \mathcal{S}_\wp J^T$. Note that $J$ is block diagonal, i.e., $\frac{\partial \varepsilon_s^i}{\partial \wp^j} = 0$, $i \neq j$.

*B. Ray intersection point covariance*

The ray intersection algorithm is based on a well-known linear algorithm where the solution minimizes the sum of squared distances from each ray to the intersection point [13]. A projection matrix $P_{r^i} = (I - \hat{r}^i(\hat{r}^i)^T)$ is defined that produces the component of a vector that is orthogonal to ray $\hat{r}^i$. That is, $d_\perp^i(X) = P_{r^i}(p^i - X)$, where $d_\perp^i(X)$ is the vector component of $(p^i - X)$ orthogonal to ray $\hat{r}^i$. The linear algorithm minimizes the sum of squared perpendicular distances, $D(X) = (\delta_\perp(X))^T \delta_\perp(X)$ where,

$$\delta_\perp(X) = \begin{bmatrix} d_\perp^0(X) \\ d_\perp^1(X) \\ \vdots \\ d_\perp^{n-1}(X) \end{bmatrix} \quad (19)$$

The solution is found by setting the derivative of $D(X)$ to zero.

$$\frac{\partial D(X)}{\partial X} = -2 \sum_i P_{r^i}(p^i - X) = 0$$

$$A = \sum_i P_{r^i}, AX = \sum_i P_{r^i} p^i, \ X = A^{-1} \sum_i P_{r^i} p^i \quad (20)$$

The solution in Eq. (20) can be seen as one that assigns the average orthogonal components of $X$ to the average orthogonal components of the ray origins.

The algorithm can be extended to take into account the error covariance of satellite pose by extracting the 2-d scalar components of the ray displacements in the coordinate system of the plane orthogonal to the ray. Define a $2 \times 3$ projection matrix $\pi_\perp^i = \begin{bmatrix} (\hat{u}^i)^T \\ (\hat{v}^i)^T \end{bmatrix}$ that maps a 3-d vector into the corresponding 2-d plane coordinates. The 2-d vector $d_\perp^i(X) = \pi_\perp^i(p^i - X)$ lies in the orthogonal plane as shown in Fig. 9.

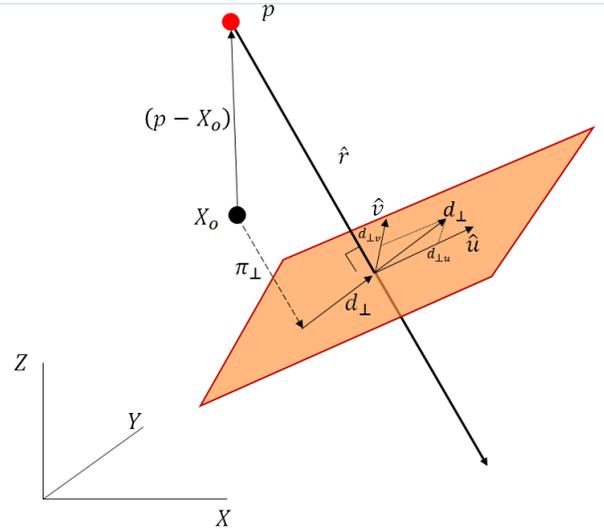

Fig. 9 The projection of a 3-d vector into 2-d coordinates of the plane orthogonal to ray $\hat{r}$.

As mentioned above, errors in satellite pose only affect the ray origin so $\frac{\partial d_\perp^i(X)}{\partial \varepsilon_u^i, \varepsilon_v^i} = \begin{bmatrix} 1 & 0 \\ 0 & 1 \end{bmatrix}$. Let the $2n \times 3$ matrix, $\Pi$, be defined as,

$$\Pi = \begin{bmatrix} \pi_\perp^0 \\ \pi_\perp^1 \\ \vdots \\ \pi_\perp^{n-1} \end{bmatrix} \quad (21)$$

Let the $3n \times 1$ matrix, $\mathcal{R}$, represent the set of vectors from the intersection point, $X$, to the ray origins, $p^i$.

$$\mathcal{R}(X) = \begin{bmatrix} (p^0 - X) \\ (p^1 - X) \\ \vdots \\ (p^{n-1} - X) \end{bmatrix} \quad (22)$$

The projection operation $\Pi \Vdash \mathcal{R}(X)$ produces a $2n \times 1$ matrix and denotes the simultaneous projection of each 3-d vector element of $\mathcal{R}(X)$, $(p^i - X)$, by the corresponding $\pi_\perp^i$ projection matrix. A weighted least squares solution can now be formulated where the scalar quantity to be minimized is,

$$\widetilde{D}(X) = \big(\Pi \Vdash \mathcal{R}(X)\big)^T (\mathcal{S}_\varepsilon)^{-1} \Pi \Vdash \mathcal{R}(X) = \\ \mathcal{R}^T(X) \Vdash \Pi^T (\mathcal{S}_\varepsilon)^{-1} \Pi \Vdash \mathcal{R}(X). \quad (23)$$

The evidence from each 2-d orthogonal displacement is weighted by the inverse ray covariance matrix so that more accurate ray positions play a greater role in the determination of the intersection point. The solution for the intersection point proceeds as follows.

$$\frac{\partial \widetilde{D}(X)}{\partial X} = -2\Pi^T (S_{\varepsilon_s})^{-1} \Pi \Vdash \mathcal{R}(X) = 0 \tag{24}$$

Define the vector of ray origins as $\mathbb{P}$, a $3n \times 1$ matrix,

$$\mathbb{P} = \begin{bmatrix} p^0 \\ p^1 \\ \vdots \\ p^{n-1} \end{bmatrix} \tag{25}$$

and $\mathbb{X}$ as a $3n \times 1$ matrix of copies of the unknown intersection point,

$$\mathbb{X} = \begin{bmatrix} X \\ X \\ \vdots \\ X \end{bmatrix} = \begin{bmatrix} I_{3\times 3} \\ I_{3\times 3} \\ \vdots \\ I_{3\times 3} \end{bmatrix} X \tag{26}$$

With these definitions, the derivative of the weighted total squared error is

$$\frac{\partial \widetilde{D}(X)}{\partial X} = -2\Pi^T (S_{\varepsilon_s})^{-1} \Pi \left( \mathbb{P} - \begin{bmatrix} I_{3\times 3} \\ I_{3\times 3} \\ \vdots \\ I_{3\times 3} \end{bmatrix} X \right) = 0 \tag{27}$$

Let $\tilde{\mathcal{A}} = \Pi^T (S_{\varepsilon_s})^{-1} \Pi$, and it follows that,

$$X = \tilde{\mathcal{A}}^{-1} \left( [[\hat{u}^0 \ \hat{v}^0] \ [\hat{u}^1 \ \hat{v}^1] \ \cdots \ [\hat{u}^{n-1} \ \hat{v}^{n-1}]] (S_{\varepsilon_s})^{-1} \begin{bmatrix} [(\hat{u}^0)^T] \\ [(\hat{v}^0)^T] \\ [(\hat{u}^1)^T] \\ [(\hat{v}^1)^T] \\ \vdots \\ [(\hat{u}^{n-1})^T] \\ [(\hat{v}^{n-1})^T] \end{bmatrix} \Vdash \begin{bmatrix} p^0 \\ p^1 \\ \vdots \\ p^{n-1} \end{bmatrix} \right) \tag{28}$$

It can be shown that the covariance of the 3-d intersection point coordinates is $\tilde{\mathcal{A}}^{-1} = P$ (see Eq. 3), and is the same as the Multi-Image Geopositioning (MIG) solution but in this case is a natural byproduct of the ray intersection solution. Eq. (28) is the key result where the computation of the multi-ray intersection point and error propagation of satellite pose covariance occurs at the same time.

To illustrate the benefit of the covariance-weighted solution an example set of 17 WorldView3 images was selected. The image set is composed of three orbital passes and the pose covariance matrix contains correlations between satellite attitude errors within each pass. The resulting $85 \times 85$ covariance matrix defines a multi-normal distribution for satellite pose errors. Samples are generated from this distribution to produce an ensemble of perturbed ray positions. The 17 rays for each sample are intersected using the original linear algorithm without covariance weighting. The scatter of the resulting 3-d intersection points for 100,000 samples is shown in Fig. 10.

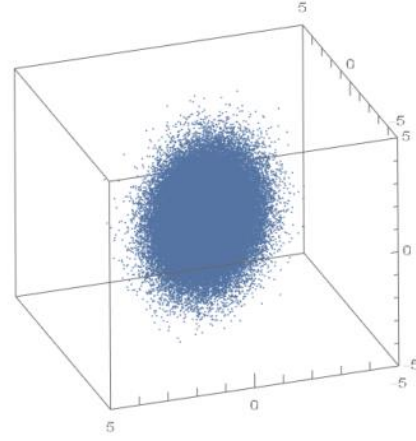

Fig. 10 The scatter of 3-d intersection points due to satellite pose error. Plot units are meters.

The covariance matrix of these samples $S_{samp}$ is compared with the covariance matrix for $X$ as determined by $\tilde{\mathcal{A}}^{-1}$. The uncertainty ellipsoid with covariance weighting has half the volume of that without covariance weighting, demonstrating much improved accuracy.

The global position of the ray intersection point is therefore significantly more accurate than its predicted position from any single image. The intersection point can be used to correct the global bias of each image by projecting the point into the image and translating the image coordinate system to align the image location of the intersection point with the location of the projected 3-d point. In addition, this alignment procedure co-registers the images to sub-pixel relative accuracy as is needed for stereo reconstruction.

## VI. LOCAL ERRORS

### A. DSM point grids

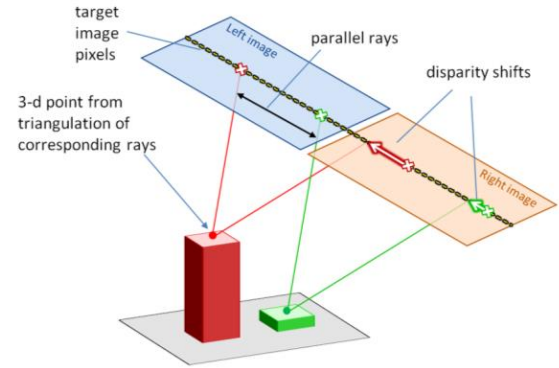

Fig. 11 Disparity shifts between the left and right image of a rectified stereo pair results in 3-d points constructed by intersecting rays.

Each stereo pair generates a disparity image found by the SGM algorithm. For details of the algorithm, see Hirschmüller [3]. The disparity values represent the difference in image locations for corresponding surface positions along rows of the epipolar-rectified left and right image. As shown in Fig. 11, surfaces closer to the sensor produce larger disparity shifts than more remote surfaces. Note that the camera rays for a given image are parallel in accordance with the affine projection model, see Eq. (8) and Fig. 6. 3-d points are found by intersecting the pair of rays defined by the disparity image.

A typical stereo pair will generate more than one million 3-d points that densely cover the observed scene surfaces.

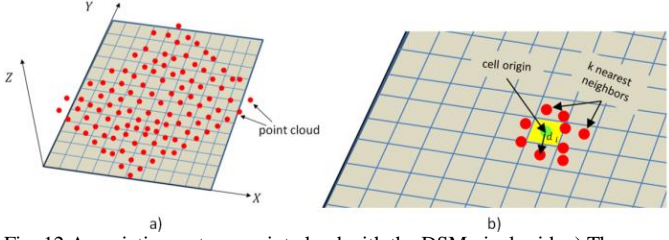

Fig. 12 Associating a stereo point cloud with the DSM pixel grid. a) The dense point cloud is not generally aligned with the DSM pixel grid. b) The elevation for each pixel cell is found by interpolating nearby values.

### B. Horizontal error

The point clouds for each stereo pair are fused together by associating the points with DSM pixel domains resulting in a set of elevation values for each pixel location as shown in Fig. 12. The stereo image pixel space and the 3-d DSM grid coordinate space will necessarily be different and so the dense point cloud generated by a stereo pair will not be aligned with DSM bins, as shown in Fig. 12a). A DSM pixel bin $(i,j)$ is associated with a set of $k$ neighbors $N_{i,j}$ that lie within a specified radius of the bin center.

That is $N_{i,j} = \{X_r | d_r \leq r, |N_{i,j}| \leq k\}$. As mentioned in the introduction, each point $X_r$ is associated with a probability $P_r$ based on the consistency in 3-d point locations between the forward and reverse order of the stereo pair. The elevation at a bin is determined by a weighted least squares algorithm where the weight $w_r$ of a point is, $w_r = \frac{P_r}{d_r}$. The point probabilities are also interpolated to assign a probability to the bin itself, $\bar{P}_q(i,j)$, with respect to a given stereo pair $q$.

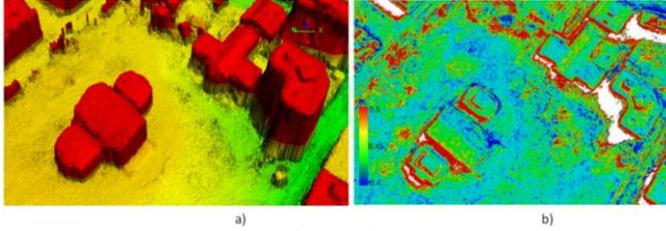

Fig. 13 The horizontal error for a DSM of Richmond, VA with 0.3 m pixel spacing and 100 fused stereo pairs. a) A region of the DSM. b) The horizontal standard deviation. Red indicates a larger standard deviation, blue smaller. The white pixels are undefined due to shadows. The peak of the distribution of $\bar{\sigma}_h$ is 0.45m, slightly larger than the grid spacing of 0.3m.

In general, the points that are assigned to a bin are scattered randomly about the bin center. Thus, the elevation assigned to the bin will not correspond to the actual world surface elevation at the bin geographic coordinates. The elevation is more accurately associated with the location of the neighborhood centroid but with a standard deviation related to the scatter of the point locations within the neighborhood. The horizontal variance at bin $(i,j)$ for stereo pair $q$ is given by,

$$\sigma_{hq}^2(i,j) = \frac{1}{\sum_r P_r(i,j)} \sum_r P_r(i,j) d_r^2(i,j) \qquad (29)$$

It is necessary to combine these horizontal variance values from each stereo pair to obtain the final variance for the fused DSM. The fused horizontal variance is given by,

$$\bar{\sigma}_h^2(i,j) = \frac{1}{\sum_q \bar{P}_q(i,j)} \sum_q \bar{P}_q(i,j) \sigma_{hq}^2(i,j) \qquad (30)$$

An example of the fused local horizontal standard deviation is shown in Fig. 13. As might be expected the standard deviation increases markedly near step boundaries at building edges. There is also somewhat higher variance on some sloped surfaces and near slope discontinuities.

### C. Vertical error

The fusion of multiple elevation values at each DSM pixel bin is achieved by forming a weighted set of elevation values that are consistent with each other. In this context a set

$$C(i,j) = \{z_q(i,j) | abs\left(z_q(i,j) - z_l(i,j)\right) < tol\} \qquad (31)$$

is consistent if all values are within a tolerance of an initial *seed* $z_l(i,j)$ used to start the formation of the set. The fusion process tries all elevation values at $(i,j)$ as seeds and selects the consensus set with the largest expected number of members $\langle N_C \rangle$, where $\langle N_C \rangle = \sum_{q \in C} P_q$ and $P_q$ is the probability of $z_q(i,j)$.

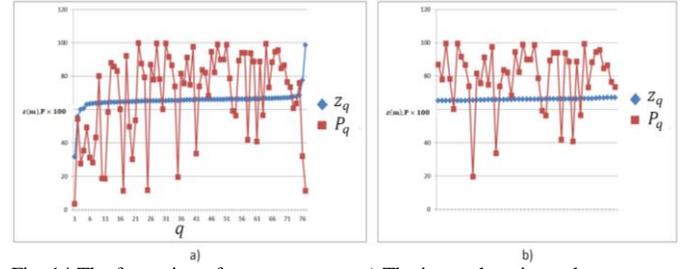

Fig. 14 The formation of consensus sets. a) The input elevation values at a pixel (u, v). b) The largest consensus set. Probability values (x100) for each elevation are shown in red.

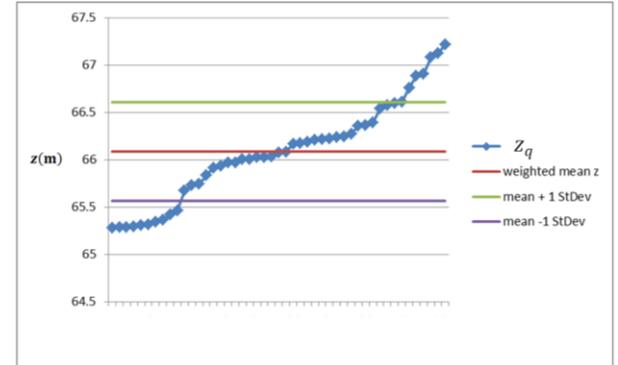

Fig. 15 The consensus set from Fig. 14b).

An example population of elevations and corresponding probabilities is shown in Fig. 14a). The elevation set is sorted to show the existence of outliers. Note that extreme values at low and high elevations have low probabilities, as well as some points with consistent values. The consensus set with the largest value of $\langle N_C \rangle$ is shown in Fig. 14b). An expanded view of the consensus set is shown in Fig. 15. The elevation value for the bin is assigned the weighted mean of the elevations,

$$\bar{z}(i,j) = \frac{1}{\sum_{q \in C} P_q} \sum_{q \in C} P_q(i,j) \, z(i,j) \qquad (32)$$

with standard deviation,

$$\sigma_z(i,j) = \sqrt{\frac{1}{\sum_{q \in C} P_q} \sum_{q \in C} P_q(i,j) \left( z(i,j) - \bar{z}(i,j) \right)^2} \quad (33)$$

The $\pm 1\sigma_z$ limits about the mean are also shown in Fig. 15.

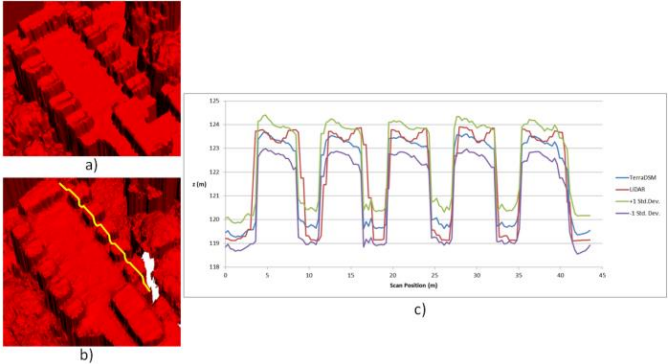

Fig. 16 An example of predicting vertical error. a) A 0.5m LiDAR scan of a building at UCSD. b) The DSM surface for the same building. The yellow curve shows the approximate scan path. c) Comparison of predicted vs. actual error, blue DSM, red LiDAR, green, purple $\pm \mathbf{1\sigma_z}$

This analysis provides a prediction of the expected local vertical errors at each pixel of the DSM. An example is shown in Fig. 16. The DSM surface is within ±1 standard deviation bounds with respect to the LiDAR except at two of the step edges. The LiDAR resolution of 0.5m and the predicted horizontal error of at least 0.3m can account for these discrepancies.

## VII. EXPERIMENTAL RESULTS

### A. Global Error

*a) Buenos Aires, Argentina*

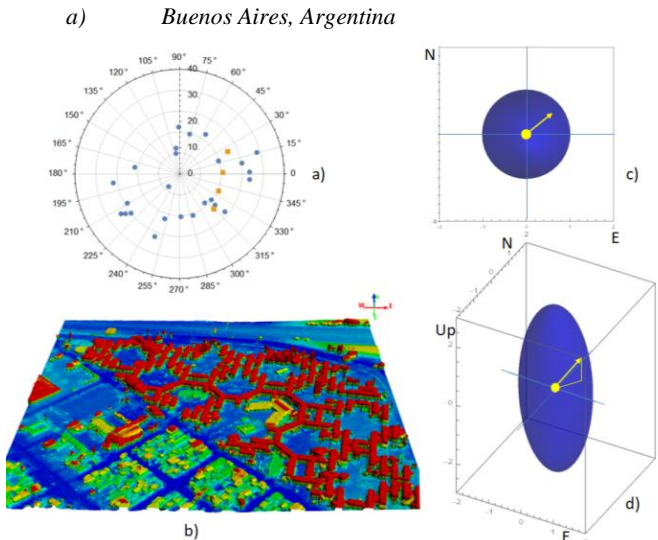

Fig. 17 Global error analysis for Buenos Aires. a) The azimuth and elevation of the satellite view directions (standard spherical coordinates). The orange squares indicate a single pass. b) The DSM for the region of interest. c), d) Views the 90% error ellipsoid in enu coordinates (m). The yellow arrow indicates geographic bias in the placement of the DSM.

A site in Buenos Aires, Argentina, (34.4894120S, 58.5859220W), is shown in Fig. 17. The feature correspondence track has 29 WorldView3 images as shown in Fig. 17a). Four of the images are collected on the same orbital pass as indicated by orange collinear points. All five of the pose components of the four images are correlated with $\rho = 0.8$. The DSM has to be translated by $t_{bias} = (0.59, 0.49, 0.98)$ m in order to align with a LiDAR DSM covering the same region. The LiDAR DSM has 0.5m spacing and the DSM Fig. 17b) has 0.3m spacing. The alignment is carried out by converting both to point clouds and minimizing the distances to the nearest points.

The effect of attitude correlation on four out of 29 images was investigated by setting $\rho = 0.0$ and reprocessing the DSM with the modified geo-registration. The registration algorithm that aligns the DSM and LiDAR point clouds has a standard deviation over repeated registration trials of about 0.1m for translation components. Thus, no significant difference is observed with $t_{bias,\rho=0} = (0.485, 0.547, 1.012)$ m. Finally, the $2n \times 2n$ covariance matrix of the ray displacements was set to the identity matrix to observe the bias without taking into account view obliquity effects. Again, no significant change in bias was observed, $t_{bias,cov.=I} = (0.441\ 0.563\ 0.980)$m. This result is to be expected since the pose variances are approximately the same for each image only differing with respect to the degree of view direction obliquity. If pose variances are identical for each image, the common variance value can be factored out of the expression for squared perpendicular error and the result reverts to ray intersection without covariance weighting.

*b) Wright Patterson AFB, Dayton, OH*

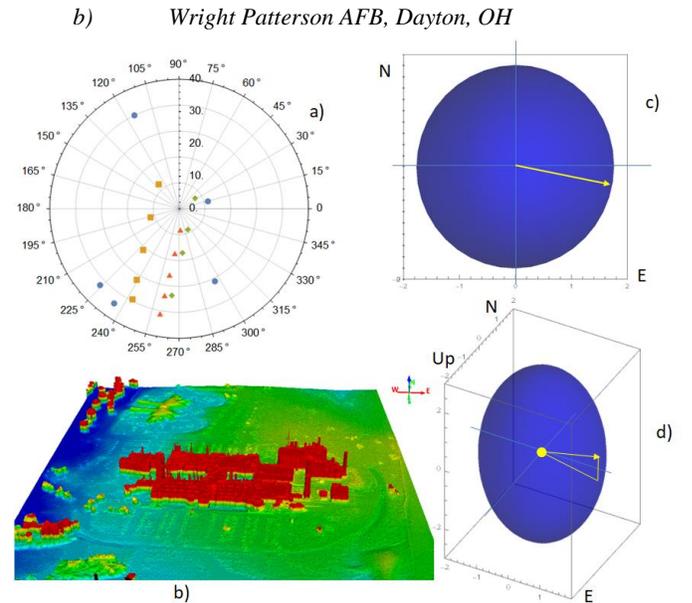

Fig. 18 Global error analysis for Wright Patterson AFB. a) The azimuth and elevation of the satellite view directions (standard spherical coordinates). There are 19 images with three single pass groups (orange, red, green). b) The DSM for the region of interest, 0.3m pixel spacing. c), d) Views of the 90% error ellipsoid in enu coordinates (m). The yellow arrow indicates geographic bias in the placement of the DSM.

The WorldView3 image data for the Wright Patterson AFB experiment is an unusual case where most of the images are in disjoint subsets that are collected within seconds of each other on single orbits. There are three single-pass groups as indicated by the orange, red and green points in Fig. 18a). There will be a high degree of correlation between the position and attitude errors of the satellite pose for each of the orbital passes. Accounting for these correlations significantly reduces the resulting horizontal (East-West) bias in the 3-d ray intersection point for the 19 image track.

The 3-d covariance matrix with pose correlation coefficient $\rho = 0.8$ is

$$P = \tilde{\mathcal{A}}^{-1} = \begin{bmatrix} 0.501598 & 0.004349 & -0.02467 \\ 0.004349 & 0.526126 & -0.02467 \\ -0.02467 & -0.02467 & 1.220778 \end{bmatrix}.$$

The required translation to align the DSM with LiDAR ground truth is $t_{bias} = 1.691, -0.385, 0.540$. The resulting bias without considering pose covariance is

$$t_{bias,cov.=I} = 2.079, -0.414, -0.153,$$

which exceeds the predicted 90% bias error ellipsoid in Fig. 18c), d). This experiment shows that correctly modeling pose correlation on single pass collection sequences decreased the horizontal error component of the bias, while it increased the vertical bias component.

propagation in the computation of the 3-d track intersection point and the resulting DSM geographic bias. The original DSM georegistration algorithm using non-linear bundle adjustment without covariance weighting had produced a bias of $t_{bias,BA} = (0.180, -1.203, 1.629)$ compared to the new $t_{bias,cov.} = (0.583, -0.696, 0.349)$ with covariance weighting and the linear least squares ray intersection algorithm. The effect of covariance weighting is very significant – the magnitude of the bias translation vector with covariance weighting is approximately one half that produced by the original bundle adjustment algorithm.

The significant difference in error covariance among the different sensor platforms emphasizes the benefit of applying error propagation in the computation of the 3-d track intersection point and the resulting DSM geographic bias.

### c) Richmond, Virginia

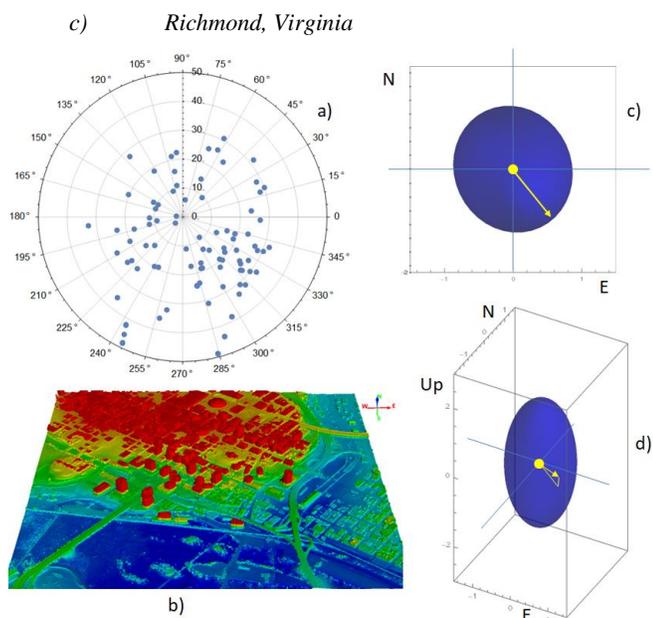

b)

Fig. 19 Global error analysis for the Richmond site. a) The azimuth and elevation of the satellite view directions (standard spherical coordinates). There are 44 images in the single correspondence track. b) The DSM for the region of interest, 0.3m pixel spacing. c), d) Views of the 90% error ellipsoid in enu coordinates (m). The yellow arrow indicates geographic bias in the placement of the DSM.

TABLE I
DISTRIBUTION OF SENSOR TYPES FOR THE RICHMOND DATASET

| Satellite Platform | Number | Position Std. Dev. | Attitude Std. Dev. |
|---|---|---|---|
| GeoEye-1 | 12 | 0.7071 | $2 \times 10^{-6}$ |
| QuickBird | 3 | 1.0 | $23.203 \times 10^{-6}$ |
| WorldView1 | 1 | 0.7071 | $3.742 \times 10^{-6}$ |
| WorldView2 | 23 | 0.7071 | $2.83 \times 10^{-6}$ |
| WorldView3 | 5 | 0.7071 | $2.83 \times 10^{-6}$ |

There is a heterogeneous mix of satellite sensor types in this example. The correspondence track has 44 images with the population of satellite platforms shown in Fig. 19. The significant difference in error covariance among the different sensor platforms illustrates the benefit of applying error

### d) Kandahar, Afghanistan

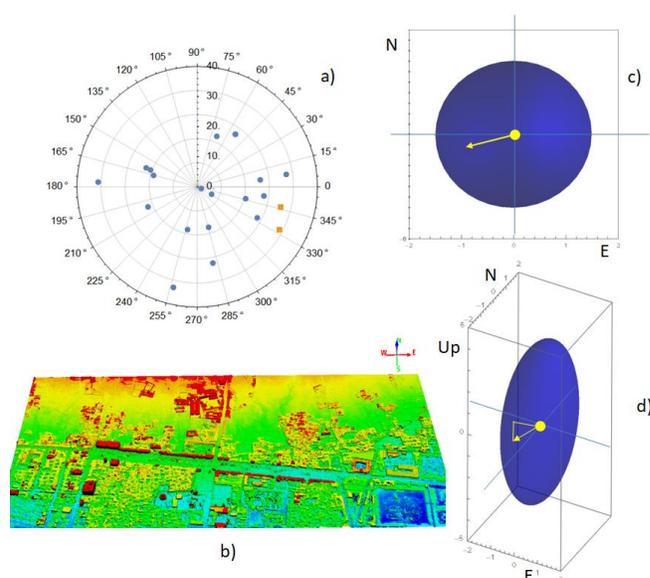

b)

Fig. 20 Global error analysis for the Kandahar, Afghanistan site. a) The azimuth and elevation of the satellite view directions (standard spherical coordinates). There are 21 images with one pair of images on the same pass. (orange). b) The DSM for the region of interest, 0.3m pixel spacing. c), d) Views of the 90% error ellipsoid in enu coordinates (m). The yellow arrow indicates geographic bias in the placement of the DSM.

TABLE II
DISTRIBUTION OF SENSOR TYPES FOR THE KANDAHAR DATASET

| Satellite Platform | Number | Position Std. Dev. | Attitude Std. Dev. |
|---|---|---|---|
| QuickBird | 2 | 1.0 | $23.203 \times 10^{-6}$ |
| WorldView1 | 6 | 0.7071 | $3.742 \times 10^{-6}$ |
| WorldView2 | 13 | 0.7071 | $2.83 \times 10^{-6}$ |

The global position bias of the Kandahar DSM, Fig. 20 is well within the 90% ellipsoid boundary. This experiment demonstrates that the pose covariance error propagation is also valid in the Eastern Hemisphere and for a range in collection times from 2009 to 2013. Again, there is a mix of platform types but with the population dominated by the WorldView2 collects, as shown in Table II.

## B. Local Error

*a)   Vertical Error*

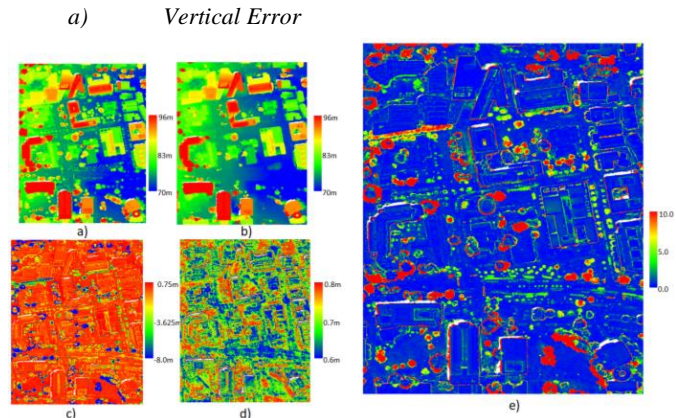

Fig. 21 Vertical error at UCSD. a) LiDAR ground truth (0.5m). b) DSM (0.3m). c) $(z - \bar{z})$, where $\bar{z}$ is the LiDAR ground truth elevation. d) The vertical error standard deviation $\sigma_z$. e) Normalized distance, $(z, \bar{z}) = \frac{|z-\bar{z}|}{\sigma_z}$.

The accuracy of vertical error prediction for a region of the University of California at San Diego DSM is shown in Fig. 21. In the experiments to follow, the DSM and the LiDAR GEOTIFFs are converted to 3-d point clouds and the closest LiDAR point to a given DSM point is considered to define the ground truth elevation. The ground truth and DSM elevations are shown as colorized images in Fig. 21a) and b). The scales indicate the range of elevations. The difference between elevations, $(z - \bar{z})$, is shown in Fig. 21c).

Note that there are areas of large negative value (blue) due to differences in vegetation between the DSM and the LiDAR. These differences are due to the relatively long time span, 2014 – 2019, of image collections used to construct the DSM and the nearly instantaneous collection time of the LiDAR. The vegetation, as well as transient vehicles, and parking lot construction, varied considerably over the five year period but all of these elevation changes are fused into a single height estimate in the final DSM. Persistent structures are accurately represented in the DSM and appear as red in Fig. 21c).

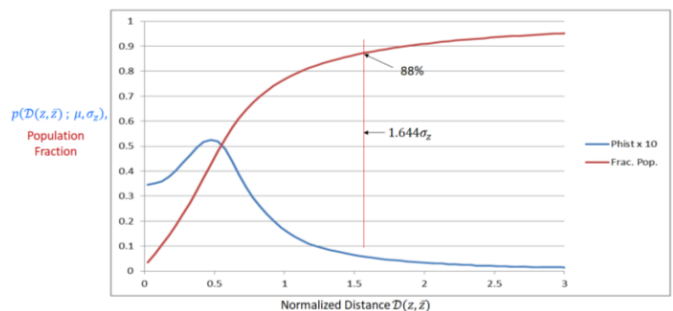

Fig. 22 The distribution of normalized distance for the region from UCSD in Fig. 21. Note that the "normalized distance" axis of the distance histogram is scaled by a factor of 10.

Other than the vegetation areas, error prediction is consistent with the observed elevation differences except at building roof boundaries. Note the fine green linear edges in Fig. 21e) exhibiting higher normalized distances. As was noted earlier, these differences may often be accounted for by the horizontal error tolerance that permits a range of ground truth positions to be matched with a single DSM elevation. This point will be examined later in the discussion on horizontal error.

Another portrayal of vertical error prediction accuracy is provided by the distribution of normalized distance over the region of interest. A histogram of the values of normalized distance $\mathcal{D}(z, \bar{z}) = \frac{|z-\bar{z}|}{\sigma_z}$ is shown in Fig. 22. In spite of the large regions of vegetation changes 88% of the pixels are within a normalized distance of 1.644 $\sigma_z$ [1]. The predicted error is somewhat conservative for smooth surfaces such as building roofs. However, unusual events such as shown in Fig. 25 indicate that conservative error bounds are often warranted for complex building structures.

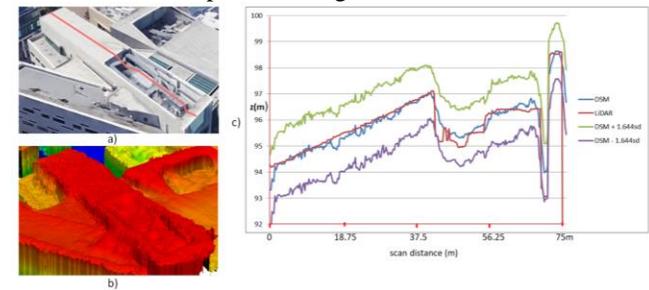

Fig. 23 A line plot comparing DSM elevations with LiDAR. a) A view of the scan path sketched on Google Earth (Copyright Google Earth) b) A view of the DSM surface for the same region. c) A line plot of the DSM values (blue) compared to LiDAR (red). Also shown are the 90% linear error bounds, green, purple ($\pm 1.64\sigma$). (Note that $\sigma$ is a function of position and so the bounds contours are not just constant offsets to the DSM elevations.)

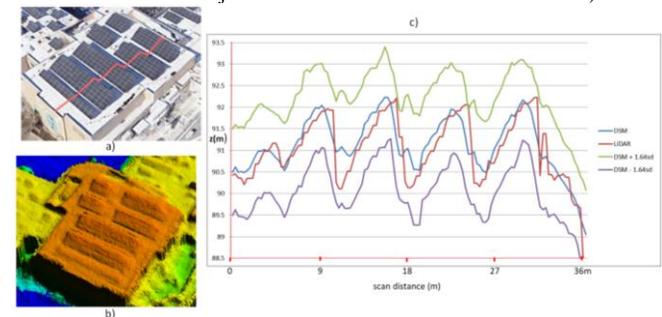

Fig. 24 A line plot comparing DSM elevations with LiDAR. a) A view of the scan path sketched on Google Earth (Copyright Google Earth) b) A view of the DSM surface for the same region. c) A line plot of the DSM values (blue) compared to LiDAR (red). Also shown are the 90% linear error bounds

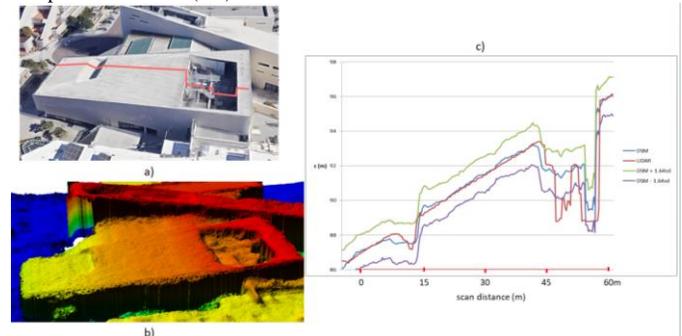

Fig. 25 A line plot comparing DSM elevations with LiDAR. a) A view of the scan path sketched on Google Earth (Copyright Google Earth) b) A view of the DSM surface for the same region. c) A line plot of the DSM values (blue) compared to LiDAR(red). Also shown are the 90% linear error bounds.

---

[1] The normalized distance that bounds 90% of normally distributed errors.

The accuracy of vertical error prediction for a sloped surface from the same region can also be conveyed by plots where the elevation values lying along a line of DSM samples is compared to LiDAR at the same sample locations. Two examples from UCSD are shown in Fig. 23 and Fig. 24 . In these two examples all the observed differences are bounded by the LE90 limits and most of the points are well within $1\sigma$ limits. There is an excursion of the LiDAR in Fig. 25 between scan distances 45m and 50m that is not within the vertical error bounds. The DSM surface was unable to recover some of the fine details of structures within the recessed cavity. In any case, 92% of the scan points are still within the LE90 limits.

*a)    Horizontal Error*

It is difficult to measure horizontal error directly for several reasons:
- the LiDAR DSM ground truth pixel spacing, 0.5m, is on the order of the horizontal error
- changes in elevation where horizontal position error might be estimated have step widths that are often larger than the horizontal error standard deviation.

Another approach is to augment vertical error bounds evaluation by examining the population of ground truth elevations within a neighborhood defined by the horizontal error CE90 radius. The closest ground truth elevation within the neighborhood to the DSM test point can be taken as the true value. The horizontal neighborhood radius $r_{h90}$ is defined as,

$$r_{h90}(i,j) = 2.146\, \bar{\sigma}_h(i,j) + \frac{1}{\sqrt{2}}\, s_{LiDAR}$$

, where $s_{LiDar}$ is the LiDAR cell spacing (0.5m). The term $\frac{1}{\sqrt{2}} s_{LiDAR}$ defines the diagonal distance to the center of a LiDAR pixel so that ground truth neighbors just touching the horizontal position uncertainty boundary are included. The term 2.146 accounts for the circular error 90% uncertainty due to normally distributed random point scatter.

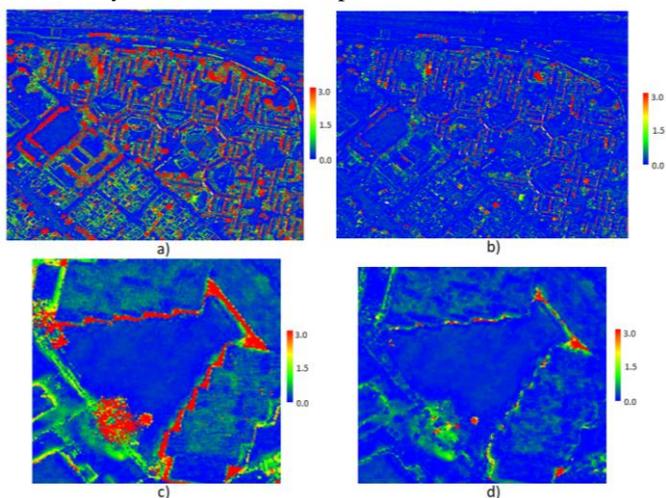

Fig. 26 Normalized distance results for Buenos Aries. a) Normalized distance with nearest LiDAR point. b) Normalized distance with closest ground truth elevation $z$ within $r_{h90}(i,j)$. c) Zoomed in view of a). d) Zoomed in view of b).

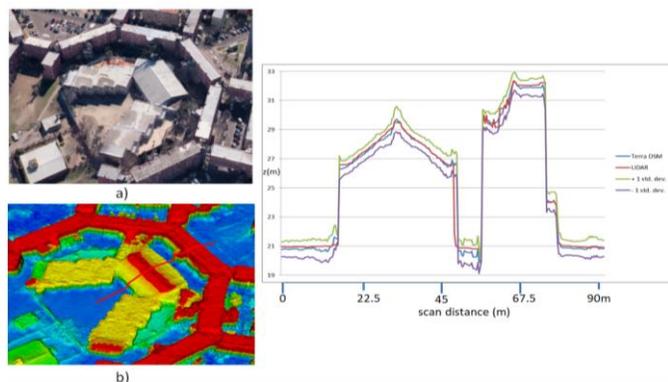

Fig. 27 A linear scan plot for Buenos Aries. a) A view of the scan region (copyright Google Earth). b) The approximate scan path on the DSM surface. c) Elevation values with $\pm 1\, \sigma_z$ bounds.

As shown in Fig. 26 the population of large normalized distance values at step boundaries is significantly reduced by introducing the neighborhood tolerance based on horizontal error. As seen in Fig. 26c) and d) the large errors at the structure boundary are mostly eliminated. The result for a linear scan across a gable roof structure adjacent to Fig. 26c), d) is shown in Fig. 27.

VIII. CONCLUSIONS AND FUTURE WORK

The error prediction results just described provide a comprehensive approach to characterizing the expected uncertainty of DSM products. It has been shown that attitude correlation for satellite image collections along the same orbital pass can have a very significant effect on the accuracy of DSM geopositioning. In the Wright Patterson AFB experiment of Fig.18, the geo-positioning bias in the Easterly direction is reduced so as to be within the 90% ellipsoidal bounds, while error prediction fails without accounting for such correlation. In the case of the Richmond, VA experiment, the application of satellite pose covariance weighting for a mix of satellite sensor types reduces the geopositioning bias vector magnitude by one half, compared to the unweighted solution, while at the same time producing a predicted 90% ellipsoid that properly bounds the actual observed bias.

The approach to local elevation error prediction enables the specification of uncertainty at every DSM sample. These predictions bound the observed error with respect to registered LiDAR within one standard deviation at most surface locations with the exception of surface step discontinuities and vegetation changes due to the different times of acquisitions of the datasets. The errors at such boundaries are significantly higher than predicted by the standard deviation computed from DSM bin elevation consensus sets and in some cases do not lie within the LE90 bounds defined by $1.64\sigma_z$. These discrepancies are due largely to errors in the horizontal location of the step boundary and to the discrete spacing (0.5m) of the LiDAR ground truth surface points. If the horizontal tolerance is relaxed to include a neighborhood of ground truth points then the elevation errors at step boundaries are dramatically reduced with most surface points lying within the predicted bounds.

The predicted error is significantly higher than observed error for smooth surfaces such as building roofs. Future investigations will consider the measured DSM surface roughness within a neighborhood around each DSM grid cell in forming a tighter bound on predicted local error over smooth surfaces. That is, the error prediction will be conditioned on local surface geometry.

A key emphasis for future experiments will focus on expanding the number and type of sites to validate both global bias terms and local error contributions to predicted geolocation uncertainty. It is also important to carefully analyze and select ground truth regions where the scene content has not changed significantly between the respective sources in order to faithfully evaluate the 3-d disparities between the LiDAR content and the DSM surfaces.

The use of satellite pose covariance weighting to improve DSM geographic placement accuracy depends on the accuracy of the specified position and attitude variances. It is possible to estimate these variances from correspondence tracks that have a small absolute geographic bias relative to the satellite pointing error. Various points on the satellite orbit will be examined to determine pose error variance estimates and their variation with respect to satellite position and attitude.